\documentclass[journal]{IEEEtran}

\ifCLASSINFOpdf
\else
\fi

\ifCLASSOPTIONcompsoc
    \usepackage[caption=false, font=normalsize, labelfont=sf, textfont=sf]{subfig}
\else
	\usepackage[caption=false, font=footnotesize]{subfig}
\fi

\usepackage{cite}
\usepackage[version=4]{mhchem}
\usepackage{graphicx}
\usepackage{amsmath,amssymb,amsfonts}
\usepackage{cleveref}
\usepackage[table,xcdraw]{xcolor}

\usepackage{algorithm}% http://ctan.org/pkg/algorithms
\usepackage{algorithmic} 

\graphicspath{{figures/}}

\begin{document}

\title{Intelligence Processing Units Accelerate Neuromorphic Learning}
% \maketitle
%%%%% Title Candidates? %%%%%   
% Accelerating Spiking Neural Network Training Using IPUs in snnTorch
%% Accelerating Spiking Neural Network Training Using Intelligent Processing Units
%% Intelligent Processing Units Accelerate Spiking Neural Net Workloads

\author{Pao-Sheng~Vincent~Sun,
		Alexander~Titterton,
		Anjlee~Gopiani,
		Tim~Santos,
		Arindam~Basu,~\IEEEmembership{Senior~Member,~IEEE,}
		Wei~D.~Lu,~\IEEEmembership{Fellow,~IEEE,}
		and~Jason~K.~Eshraghian,~\IEEEmembership{Member,~IEEE}\\%
		
		Code: https://github.com/vinniesun/snntorch-ipu
        
\thanks{P.~V.~Sun and A.~Basu are with the Department of Electrical Engineering, City University of Hong Kong, Hong Kong, SAR.}% <-this % stops a space
\thanks{A. Titterton, A. Gopiani and T. Santos are with Graphcore, Bristol, UK.}%
\thanks{W.~D.~Lu is with the Department of Electrical Engineering and Computer Science, University of Michigan, Ann Arbor, Michigan 48109, USA.}%
\thanks{J.~K.~Eshraghian is with the Department of Electrical and Computer Engineering, University of California, Santa Cruz, CA 95064, USA. He was previously with the Department of Electrical Engineering and Computer Science, University of Michigan, Ann Arbor, Michigan 48109, USA. (jeshragh@ucsc.edu)}
% \thanks{Manuscript received November 16, 2022.}
}%

% The paper headers
% \markboth{IEEE Internet of Things Journal}%
% \markboth{}
% {Sun \MakeLowercase{\textit{et al.}}: Accelerating Spiking Neural Network Training Using Intelligence Processing Units}

\maketitle

% Alt title: Accelerating Sparse Gradient Descent in snnTorch Using Intelligent Processing Units

\begin{abstract}

% it becomes more expensive than non-spiking networks when trained on modern graphics processing units (GPUs) because sequential processing of the temporal dimension must be accounted for. 
% Modern GPUs are not optimized for the sequential nature of temporal workloads as memory complexity scales with sequence length when training networks using backpropagation through time (BPTT).

% Domain-specific neuromorphic hardware typically relies on off-chip optimization, or is otherwise limited to rudimentary learning rules.
Spiking neural networks (SNNs) have achieved orders of magnitude improvement in terms of energy consumption and latency when performing inference with deep learning workloads. 
Error backpropagation is presently regarded as the most effective method for training SNNs, but in a twist of irony, when training on modern graphics processing units (GPUs) this becomes more expensive than non-spiking networks. The emergence of Graphcore's Intelligence Processing Units (IPUs) balances the parallelized nature of deep learning workloads with the sequential, reusable, and sparsified nature of operations prevalent when training SNNs. IPUs adopt multi-instruction multi-data (MIMD) parallelism by running individual processing threads on smaller data blocks, which is a natural fit for the sequential, non-vectorized steps required to solve spiking neuron dynamical state equations. 
We present an IPU-optimized release of our custom SNN Python package, \textit{snnTorch}, which exploits fine-grained parallelism by utilizing low-level, pre-compiled custom operations to accelerate irregular and sparse data access patterns that are characteristic of training SNN workloads. We provide a rigorous performance assessment across a suite of commonly used spiking neuron models, and propose methods to further reduce training run-time via half-precision training. By amortizing the cost of sequential processing into vectorizable population codes, we ultimately demonstrate the potential for integrating domain-specific accelerators with the next generation of neural networks.
%We show speed improvements of \textcolor{red}{4.09$\times$ to 7.75$\times$ over commercial GPUs for dense and convolutional spiking network architectures, demonstrating the potential for integrating modern deep learning accelerators with the next generation of neural networks. }

%\url{https://github.com/jeshraghian/QSNNs}}.

\end{abstract}

\IEEEpeerreviewmaketitle

\section{Introduction}
\IEEEPARstart{R}{epurposing} GPUs from graphics rendering to training deep neural networks has effectively shaped an entire decade of advances in artificial intelligence (AI) \cite{chellapilla2006high, oh2004gpu, fatahalian2004understanding, ciresan2011flexible, krizhevsky2012imagenet}. This can be attributed to the numerous processor cores in GPUs that enable high parallelization of easily decomposable instructions, which are essential for the large number of matrix operations that take place in neural networks.

But a significant discrepancy arises: the cost of training deep learning algorithms in data centers sits between 100s and 100,000s of watts, whereas brain-driven cognition is bounded to approximately 10-20~W. This gap in performance has driven the neuromorphic engineering community to explore new algorithms, architectures, circuits, and devices that apply principles of neural processing to modern neural networks \cite{neftci2019surrogate, sussillo2009generating, jo2010nanoscale, hochstetter2021avalanches, maass1997networks, diehl2015unsupervised}. Spiking neurons transmit information in voltage bursts known as `action potentials', which are characterized as discrete events in many neural coding studies. As such, SNNs distribute information over time, where most neurons are dormant at any instantaneous moment in time. This reduces memory access frequency, which is one of the dominant costs in deep learning workloads \cite{brette2012simulating, azghadi2020hardware,fidjeland2010accelerated, eshraghian2022memristor, sze2017efficient}.

% can be trained to be in a resting state for most 

%% ex abstract
% \textcolor{red}{NOTE: CONTINUE FROM HERE} 

% To keep pace with exponentially growing deep learning workloads, custom neural network accelerators are increasing in use for processing large-scale deep learning models. At the algorithmic layer, a potential way forward to improving the efficiency of DL algorithms are spiking neural networks (SNNs), which draw inspiration from the theoretical underpinnings of biological neuron models.

% SNNs shine in this regard. Since SNNs are designed to be a better imitator of the way human brain works, SNNs, compared to other neural networks of similar depth and neuron counts, consume significantly less power and with the use of specialized hardware, such as Neuromorphic chips [13], [14], [15],even less power is consumed during training and inferencing. 

% SNNs distribute information over time, where data is encoded in the timing or frequency of spiking activity. In general, neurons are at rest by default, and only excited at irregular intervals which, in theory, reduces memory access frequency. This optimizes power efficiency by using advances at the algorithmic layer. 

% Modern GPUs are not optimized for the sequential nature of temporal workloads as memory complexity scales with sequence length when training networks using backpropagation through time (BPTT).

% Domain-specific neuromorphic hardware typically relies on off-chip optimization, or is otherwise limited to rudimentary learning rules.
When it comes to training via gradient descent, there are next to no accelerators optimized for SNN workloads. The most common uses for neuromorphic hardware are: 1) inference using fixed weights where training takes place `offline', or 2) online learning using simple plasticity rules, such as spike time-dependent plasticity (STDP). If SNNs are so efficient, why are there no accelerators that can can perform backpropagation on SNN models? While feedforward computation is cheap, in a twist of irony, gradient-based optimization of SNNs is less efficient than its non-spiking counterpart. There are several reasons for this drop in efficiency: 1) the time complexity of backpropagation through time (BPTT) means each time step instantiates an additional neural network. Memory usage scales linearly with time; 2) biological neurons are more complex than artificial neurons, and 3) the non-differentiability of spikes means that a direct application of automatic differentiation is incompatible with SNNs. In effect, GPUs and many accelerators have not been optimized for sequential instruction sets that are required by spiking neurons: multiply-accumulate $\rightarrow$ state update $\rightarrow$ thresholding $\rightarrow$ surrogate gradient calculations.

\begin{figure*}[!t]
    \centering
    \includegraphics{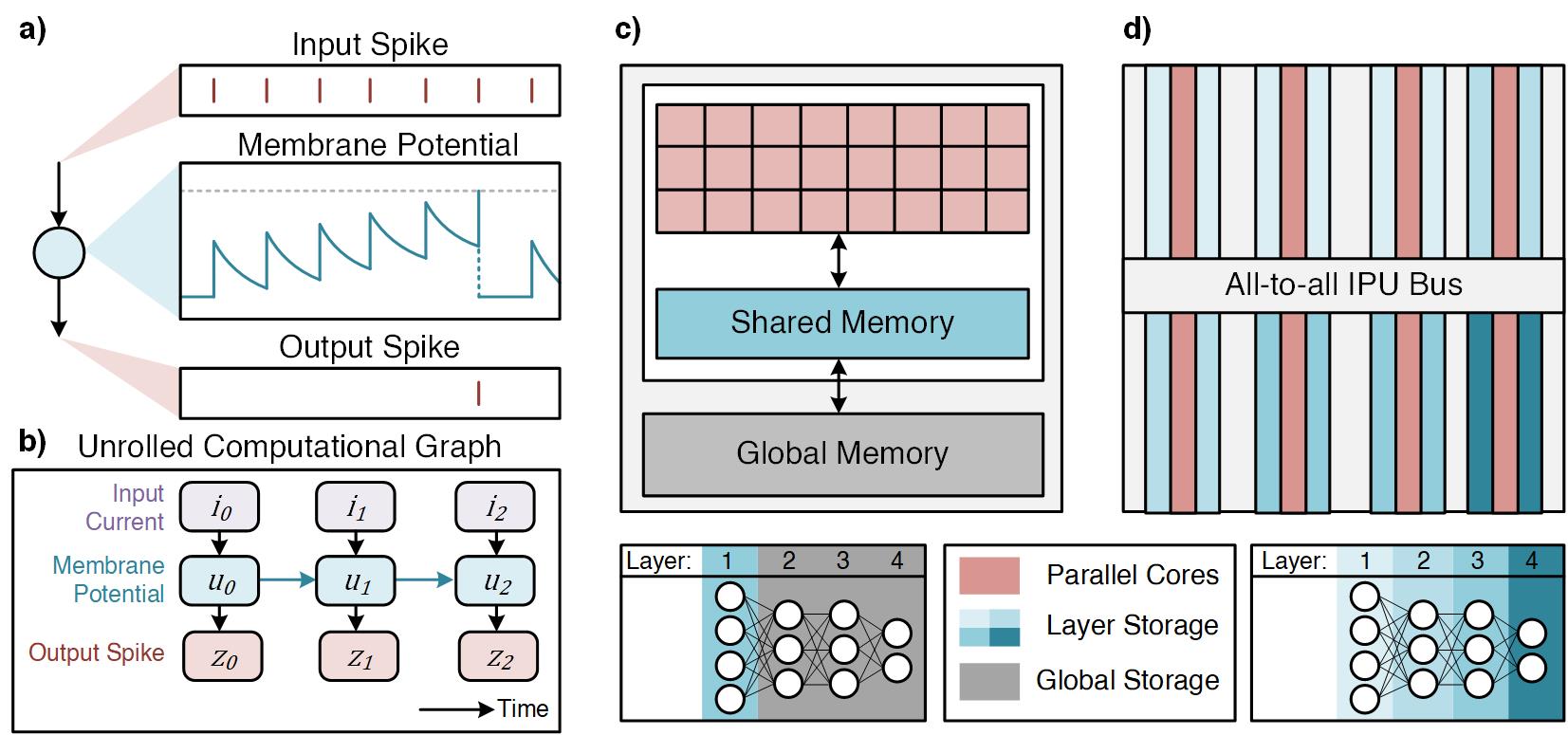}
    \caption{Mapping neural networks to hardware. (a) Dynamics of a spiking neuron model. (b) The computational graph of the neuron is unrolled over time to enable compatibility with the BPTT training algorithm. (c) SIMD/SIMT is used to perform parallel computations for one layer at a time in GPUs. (d) MIMD is used to distribute layers and custom activations, such as spiking dynamics, across IPU cores to improve concurrency. Layer-to-memory maps are color-coded.}
    \label{fig:1}
\end{figure*}

While the current market of accelerators are tailored to conventional DL workloads, this paper seeks to explore the use of accelerators that are better tailored for the types of operations that are characteristic of SNNs \cite{jouppi2017datacenter, elbtity2022aptpu, abts2020think}.
In particular, Intelligence Processing Units (IPU, \textit{Graphcore}) include a feature set that are a natural fit for training SNNs via error backpropagation. By coupling highly-parallel multi-instruction multi-data (MIMD) processing to sparse, spike-based tensors, we take a stride towards extracting the benefits from DL accelerators and porting them to neuromorphic algorithms.

% Most neuromorphic hardware relies on off-chip optimization or only integrate rudimentary local learning rules. Graphcore’s Intelligence Processing Unit (IPU) is a promising opportunity to integrate highly-parallel multi-instruction multi-data (MIMD) processing to spiking neurons, together with compiling spike-based tensors into computational graph. 

The contributions of this paper are as follows:
\begin{itemize}
    \item Our SNN Python framework, \textit{snnTorch}, is released for IPU compatibility using low-level, pre-compiled custom operations; % rather than going through numerous layers of abstraction;
    \item A variety of benchmarks are assessed to demonstrate up to 21.3$\times$ peak improvement in throughput over NVIDIA A100 GPUs when training SNNs;
    \item A series of corner cases are identified where GPUs converge to accelerator performance in recurrent SNNs;
    % 10$\times$ improvements in mixed-precision training throughput across a variety of neuron models;
    \item In much the way that brains distribute firing rates across pools of neurons, we demonstrate how the use of population codes can significantly accelerate the training process.
\end{itemize}

This paper presents the first analysis of the suitability and performance of IPUs in handling neuromorphic workloads when trained using approaches prevalent in deep learning.

% \textcolor{red}{The contributions of this paper as follows: * snnTorch updated with low-level pre-compiled custom operations; }
% Our custom Python package, snnTorch, is optimized for Graphcore’s IPUs by compiling key functions such as Heaviside function, to directly utilize the IPU’s hardware instead of going through numerous layers of abstraction, lowering the performance of the training process, thereby demonstrating the promising fit of modern DL accelerators with the next generation of neural networks. 
% To test the performance of the IPU against commercial GPUs that is using snnTorch as the main toolbox for training, we measured the training throughput is and compared the result to see how much improvement is gained through using specialized hardware like the IPU.

\section{Background}\label{background} 
\subsection{Spiking Neural Networks}
The adoption of deep learning-based techniques to training SNNs can be dated back to 2002, when Bohte \textit{et al.} treated the firing time of a spiking neuron as a trainable, regression problem \cite{bohte2002error}. Since the advent of CUDA-accelerated Python packages with built-in automatic differentiation (autodifferentiation) engines (e.g., PyTorch \cite{paszke2019pytorch}, Tensorflow \cite{abadi2016tensorflow}, JAX \cite{frostig2018compiling}), the broader approach in recent years has been to apply a generalized backpropagation algorithm to an unrolled computational graph of spiking neurons (\Cref{fig:1}(a-b)) \cite{hunsberger2015spiking, shrestha2018slayer, bellec2018long, henkes2022spiking, esser2016convolutional, huh2017gradient}. BPTT adopts techniques used to train recurrent neural networks (RNNs), where sequences are instead interpreted as discrete time-steps of finite duration \cite{pineda1987generalization, werbos1990backpropagation}.

While a variety of detailed models are used to accurately emulate biological neurons, the simplest models are more commonly used in large-scale simulations. This can be attributed to several reasons: i) calculating the solution is computationally cheap, ii) simplifying an action potential to a single-bit event promotes sparse computations, and iii) applying gradient descent to stiff equations (e.g., with sharp bifurcations) can lead to instability when training a network.

%%%%%% consider changing this
SNNs adopt the same topological structure as non-spiking networks. The main difference is that artificial neuron models are swapped out for time-varying spiking neurons. Time-evolution is modeled in a sequential structure. Specific details regarding the types of neuron models used are provided in the experimental results (\Cref{sec:exp}).

\subsection{Neuromorphic Processors}
The neuromodulatory processes in the brain that leverage spikes to promote learning remain somewhat shrouded in mystery, which has inspired the development of several research-based neuromorphic processors. Two prolific examples include Loihi developed by Intel Labs \cite{davies2018loihi, orchard2021efficient}, and SpiNNaker initiated at the University of Manchester \cite{khan2008spinnaker, furber2014spinnaker}, both of which have roused neuromorphic research ecosystems where hardware access is offered both remotely and physically to the broader research community. While such neuromorphic processors remain to be optimized for gradient-based learning, they have incited much interest in how neurobiological processes can be modelled in-silico. These processors allow users to explore how programmable learning rules can modulate plastic synapses. 

The push towards data-driven benchmarks from deep learning has led to the adoption of gradient-based learning rules to be used with SNNs, which is far better suited for non-convex optimization, but demand far more computational resources when compared to biophysically motivated learning rules. Training SNNs via gradient descent compounds upon several challenges:

\begin{itemize}
    \item \textbf{Temporal Credit Assignment:} The BPTT learning rule requires storage of all gradients over time, where memory complexity scales with $\mathcal{O}(nT)$ where $n$ is the number of neurons and $T$ is the duration of time.
    \item \textbf{Weight Credit Assignment:} Routing gradients from the network's output back to plastic synapses requires the data path of the forward operation to be stored. The gradient of every synapse has an independent pathway, which scales the cost of communicating gradients to apply weight updates.
    \item \textbf{Non-differentiable operations:} In leaky integrate-and-fire neuron models, a hard threshold is often applied to the membrane potential to elicit a voltage spike at the axon. This is a non-differentiable operation, and thus, incompatible with gradient descent.
\end{itemize}

\subsubsection{Temporal Credit Assignment}
The temporal credit assignment problem can be addressed by adopting real-time recurrent learning (RTRL) techniques, to avoid having to store gradients in time \cite{williams1989learning}. The cost of doing so is that memory complexity now scales with $\mathcal{O}(n^3)$, where the cubic term discourages broad adoption in large-scale networks. Approximations of RTRL recently inspired the development of a lightweight SNN training accelerator for fixed, dense architectures \cite{frenkel2022reckon, bellec2020solution}.

\subsubsection{Non-differentiable Operations}
Surrogate gradient descent is used to bypass non-differentiable operators, where the final calculated gradients are a sufficient approximation \cite{neftci2019surrogate, zenke2021remarkable}. This adds to computational cost, as analytical methods to computing derivatives (e.g., dual numbers \cite{griewank2008evaluating}) must be supplemented with manually-determined heuristics (surrogate gradients); i.e., training SNNs via surrogate gradients is not as modular as non-spiking networks.

\subsubsection{Low-cost Inference}
The high cost of training SNNs using non-local learning algorithms can be partially offset by the incredibly cheap cost of using SNNs in solely feedforward operations. It has been shown that SNNs can offer 2--3$\times$ orders of magnitude improvement over non-spiking alternatives \cite{azghadi2020hardware}. In general, this motivates offline training of SNNs typically using GPUs, where deployment can take place on low-power SNN accelerators. 
%\textit{Stewart et al.} successfully applied offline training of an SNN coupled with online few-shot learning that harnesses local learning for fine-tuning \cite{stewart2020online}. 
Several recent studies have leveraged programmable microcode of neuromorphic research processors to adopt BPTT variants on a single chip \cite{renner2021backpropagation, tang2021biograd}. At present, these methods are constrained to fixed neuron models and network architectures, not yet generalized to convolutional networks. Despite these limitations, such methods offer a promising alternative for online deployment of BPTT-like training of SNNs to what we propose here. Rather than taking BPTT to processors optimized for SNNs, we use IPUs to compile and train SNNs using accelerators optimized for backpropagation.

% Several neuromorphic chips have been designed for ultra-efficient training and inference of SNNs. In general, training adds overhead its driven most of the backprop methods to offline training for weights to be ported onto neuromorphic hardware. Some of the well-known neuromorphic chips includes the following:
% •	Intel’s Loihi [15]. The Loihi enables several local learning rules (IEEE Micro) such as STDP, which is typically only used for simple datasets. Offline training is delegated to using SLAYER, a framework for backpropagation based SNN learning [9].
% •	IBM’s TrueNorth [17]. TrueNorth was designed to for low power neural network based application.
% •	SpiNNaker, a chip that is based on SNN, is based on parallel, multicore computer architecture.
% There are several other designs that utilize lightweight variants of backpropagation have been proposed that show promise for physical integration. 

\subsection{Intelligence Processing Units}

% offer a simplified block diagram in Figure 1.1. Its design aim is the efficient
% execution of fine-grained operations across a relatively large number of parallel threads. This means that the IPU, unlike other massively parallel architectures (e.g., the GPU) adapts well to fine-grained, irregular computation that
% exhibits irregular data accesses. The IPU offers true MIMD (Multiple Instruction, Multiple Data) parallelism and has distributed, local memory as its only
% form of memory on the device.

IPUs are designed to facilitate deep learning workloads by processing fine-grained operations across a large number of parallel threads. The ability to process individual threads on sub-blocks offers a two-fold benefit on SNN workloads over single-instruction-multiple-data/thread (SIMD/SIMT) GPUs: i) instructions from different network layers can be concurrently processed, where the constraints of contiguous vectorized data is no longer a performance bottleneck, and ii) MIMD processing can accelerate applications with irregular and sparse data access without incurring performance degradation. This is optimal for spike-based workloads which include additional processing overhead in computing the state-driven dynamics of spiking neuron models (\Cref{fig:1}(c-d)).

% SNNs consist of a more diverse set of computations than conventional deep learning algorithms. Each spiking neuron follows dynamical state trajectories which enable SNNs to carry their ability to represent rich temporal features, but this comes at the cost of calling for multiple instructions, which is MIMD by nature, and the IPU is capable of running processing threads in parallel MIMD form.

% based applications [3], [4]. Traditionally, ML/DL models are trained with CPUs and GPUs, which are scalar and vector processor, respectively, Graphcore’s IPU opts for an alternative design, which is graph focused. In this paper, the network of focus is the SNN. SNN consist of a larger set of instructions than conventional non-spiking networks. 

Each IPU Mk2 core consists of 1,472 high performance processor cores, where each processor core and a locally accessible in-processor memory unit form a tile. The IPU tile consists of one computing core and 624~KB of local memory. Each core contains six processor threads, totaling 8,832 processor threads when operating in parallel. This amounts to a total of roughly 900~MB of memory and 250~TeraFLOPS of compute for the Mk2 GC200 IPU hardware which ran the experiments on this paper. 
%Each core contains six processor threads, totaling 8,832 processor threads when operating in parallel. The IPU processor core consists of one computing core and 256~KiB of local memory. This amounts to a total of roughly 300MB of memory. 
Each core is connected directly to the IPU-Exchange, which is capable of transferring 8~TBps of data between IPU tiles. There is no global memory, and specialized hardware is incorporated for common neural network operations, such as convolutions and matrix multiplications.

IPUs follow a graph processing pipeline where programs are compiled into a logical execution graph. This graph is composed of alternating state and computation vertices. Each vertex consists of machine instructions that can execute in parallel, provided they write to independent parts of a tensor. Upon completion of a compute step, data is exchanged between tiles as part of the exchange phase of the bulk synchronous parallel (BSP) execution model.

%data is routed between cores to ensure consistency of the system state. 

Adopting this BSP execution model benefits bandwidth-limited neural network, as overlapping memory-bound computation and communication can lead to bandwidth contention and data collision \cite{cormen1996bridging, langguth2018memory}. BSP eliminates the need for message buffers and global memory, though as a result, all inter-core communication must be planned during model compilation \cite{burchard2021ipug}. In practice, once the model has been compiled once, it can be cached and subsequently reused.

% All codelets must be completed before moving onto the next vertex of the computation graph. 

% The rationale for this structure is that due to bandwidth contention, overlapping memory-bound computation and communication is difficult and sometimes impossible [25]. Furthermore, it provides a clear computation structure and obviates the need for message buffers and thus additional memory on the chip, making communication very efficient. On the other hand, this brings about that all communication must be planned at compile time. 

%%%%%%

\subsection{snnTorch}
A variety of gradient-based SNN libraries have been open-sourced, most of which are written in Python for syntactical ease, and several of which are built on top of commonplace deep learning packages \cite{shrestha2018slayer, knight2021pygenn, eshraghian2021training, SpikingJelly, hazan2018bindsnet, norse2021}. Most approaches compose primitive functions together wrapped as a spiking neuron node, where gradients are analytically calculated using reverse autodifferentiation in the backend. As spikes are represented as discontinuous voltage bursts, they are non-differentiable. PyTorch allows users to override gradients with custom functions, and so has become a common backend for the implementation of surrogate gradient descent in SNNs \cite{neftci2019surrogate, zenke2021remarkable}.

\textit{snnTorch} is adopted as the toolbox because it is: i) designed with PyTorch as its backbone, so pre-existing interfaces can be used to lower composable PyTorch functions into IPUs, ii) several features are unique to \textit{snnTorch} in the context of gradient-based learning, such as using population-based embeddings to accelerate the training process, and iii) quantization-aware training has been integrated into the state-space of spiking neuron models, which can be used in mixed- and low-precision accelerators.

Several alternative options are available for accelerating SNNs using CUDA-based libraries. \textit{SpikingJelly} provides a CuPy backend \cite{SpikingJelly}, \textit{GeNN} uses CUDA-generated code to implement an approximate form of BPTT \cite{knight2022efficient, bellec2020solution}, and \textit{lava-dl} incorporates the most commonly used functions/neurons as optimized CUDA code, while other libraries mostly depend on the deep learning package's CUDA acceleration.

To summarize, prior approaches for faster gradient-based training of SNNs include:
\begin{itemize}
    \item Utilizing microcode to enable neuromorphic processors to track gradients,
    \item Using custom CUDA backends to accelerate SNNs on GPUs, and
    \item Using pre-existing interfaces to CUDA via pre-existing deep learning libraries (e.g., PyTorch).
\end{itemize}

The first option is burdened with instruction set-level definitions that must be tailored for a given network architecture, and the latter two are limited by SIMD/SIMT processing. We take a wholly different approach by adapting Python-level SNN descriptions that leverage low-level, pre-compiled operations customized to an IPU accelerator harnessing MIMD architectures. %While it is not silicon-optimized for spiking neurons, 
This approach to distributed memory amongst IPU cores can be used to reduce data movement, thus amortizing the costs of weight and temporal credit assignment.

% \begin{figure} 
%     \centering
%   \subfloat[\label{1a}]{%
%       \includegraphics[width=0.45\linewidth]{Figures/fig2a.jpg}}
%     \hfill
%   \subfloat[\label{1b}]{%
%         \includegraphics[width=0.45\linewidth]{Figures/fig2b'.jpg}}
% %     \\
% %   \subfloat[c\label{1c}]{%
% %         \includegraphics[width=0.45\linewidth]{example-image}}
% %     \hfill
% %   \subfloat[d\label{1d}]{%
% %         \includegraphics[width=0.45\linewidth]{example-image}}
%   \caption{(a), (b) Some examples from CIFAR-10 \cite{4}. The objects in     
%         single-label images are usually roughly aligned.(c),(d) However, the 
%         assumption of object alignment is not valid for multi-label
%         images. Also note the partial visibility and occlusion
%         between objects in the multi-label images.}
%   \label{fig1} 
% \end{figure}

\begin{figure}[!t]
    \centering
    \includegraphics[width=0.45\textwidth]{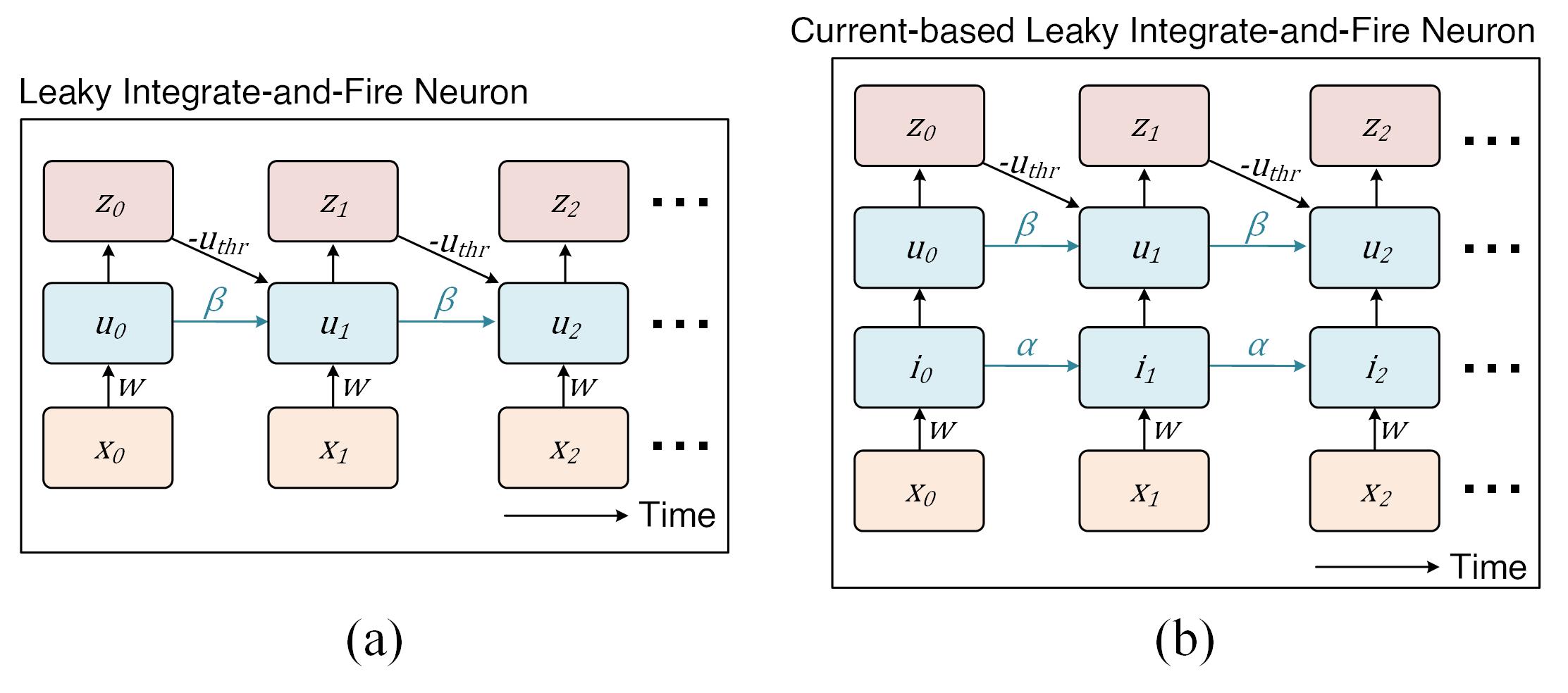}
    \caption{Unrolled computational graph of (a) Leaky Integrate-and-Fire neuron, and (b) Current-based leaky integrate-and-fire neuron.}
    \label{fig:graph}
\end{figure}

\section{Methods}\label{methods}
\subsection{Neuron Models}
\subsubsection{Leaky Integrate-and-Fire Neuron}
The dynamics of a leaky integrator neuron are as follows \cite{dayan2005theoretical, lapicque1907louis}:

\begin{equation}
\label{eq:1}
    \tau\frac{du}{dt} = -u + ir,
\end{equation}

\noindent where $u$ is the membrane potential of the neuron, $i$ is the current injection to the neuron, $r$ is the equivalent resistance of the ion channels of the neuron, $\tau=rc$ is the time constant of the neuron, where $c$ is the capacitance of the passive membrane. \Cref{eq:1} can be solved using the forward Euler method:

\begin{equation}
\label{eq:euler}
    u_t=\beta u_{t-1} + (1-\beta)i_t,
\end{equation}

\noindent where $\beta = e^{-1/\tau}$ is the inverse time constant of the neuron membrane potential, and the subscript $t$ refers to time. When the membrane potential exceeds the threshold $u_{\rm thr}$, an output spike is generated:

\begin{equation} \label{eq:spike}
    z_t =
    \begin{cases}
      1,  & \text{\rm if $u_t > u_{\rm thr}$} \\ 
      0, & \text{otherwise.} \\
    \end{cases}  
\end{equation}

To introduce learnable parameters, the current injection term is replaced with a weighted input $(1-\beta)i \leftarrow wx$. For notational brevity, the contribution of a single weighted input is used:

\begin{equation} \label{eq:lif}
    u_t=\beta u_{t-1} + wx_t - z_{t-1}u_{\rm thr},
\end{equation}

\noindent The final term introduces a reset mechanism to the neuron. The unrolled computational graph depicting the operation of the neuron is shown in \Cref{fig:graph}(a).

\subsubsection{Current-based Leaky Integrate-and-Fire Neuron}
If the leaky integrate and fire can be thought of as a low-pass filter, the current-based method can be thought of as a pair of low-pass filters. The input synaptic current is modeled as an AMPA-receptor with a rapid rise time and gradual decay, which then modulates the membrane potential of the neuron:

\begin{equation} \label{eq:cuba1}
    i_t=\alpha i_{t-1} + wx_t,
\end{equation}
\begin{equation}\label{eq:cuba2}
    u_t=\beta u_{t-1} + i_t - z_{t-1}u_{\rm thr},
\end{equation}

\noindent where $\alpha= e^{-1/\tau_{\rm syn}}$ is the inverse time constant of the synaptic current, and $\tau_{\rm syn}$ is the equivalent time constant of the synaptic current in an analogous way to $\tau$, with the computational graph illustrated in \Cref{fig:graph}(b).

\subsubsection{Recurrent Spiking Neurons}
Both of the above neuron types can be adapted to include explicit recurrent connections. The output spikes are weighted and appended to the input. Formally, a recurrent leaky integrate-and-fire neuron is represented by:

\begin{equation} \label{eq:rlif}
    u_t=\beta u_{t-1} + wx_t + z_{t-1}(v-u_{\rm thr}),
\end{equation}

% \noindent and the current-based model modifies (5) to:

% \begin{equation} \label{eq:cuba1}
%     i_t=\alpha i_{t-1} + wx_t + vz_{t-1},
% \end{equation}

\noindent where $v$ is the recurrent weight.

\subsection{Custom Operations on IPUs} \label{sec:custom}

\begin{algorithm}[tb] \label{alg}
    \caption{Using Custom Operations with IPU}
    \label{alg:tha}
\begin{algorithmic}
    \REQUIRE Custom operation defined in C++
    \REQUIRE Makefile used to generate the Shared Object% for Python
    \REQUIRE Custom operation's shared object% to be loaded in Python
    % \REQUIRE ctypes: The Python library used to load the custom operation's shared object
    \STATE Define custom operation in a C++ file
    \STATE Use the Makefile to generate the shared object
    \STATE Load the Custom Operation's Shared Object. % in the 
\end{algorithmic}
\end{algorithm}

\begin{figure*}[!t]
    \centering
    \includegraphics{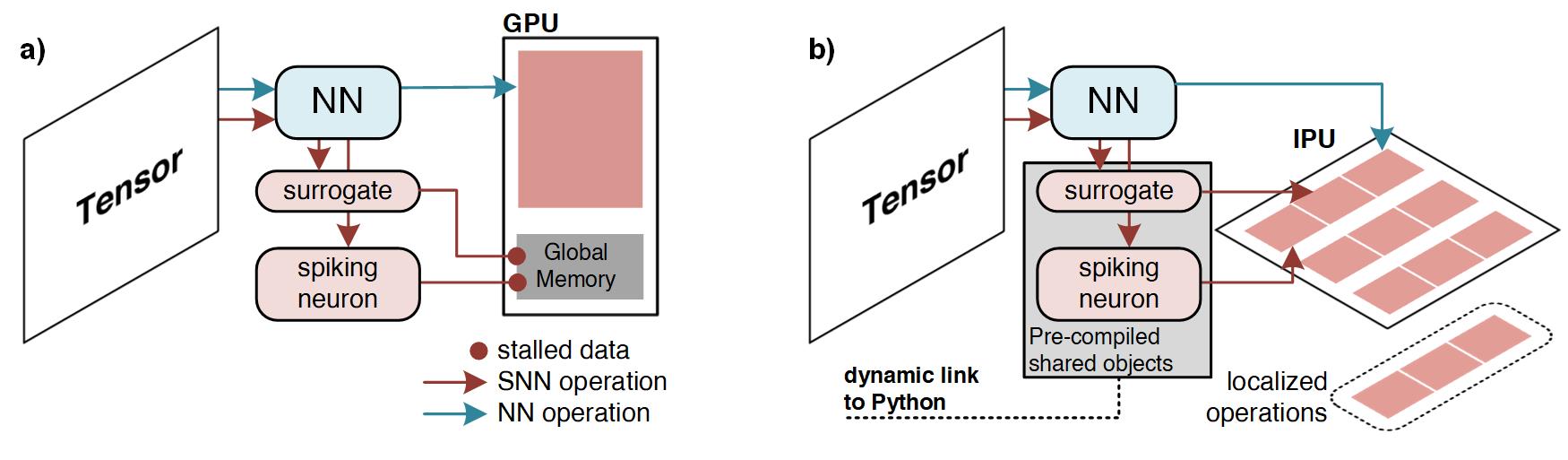}
    \caption{Data path of input tensors on GPU and IPU. (a) GPU: One instruction is applied to all elements of an input tensor while spiking neuron state and surrogate gradient computations are stalled in the instruction pipeline. (b) IPU: Spiking neuron state and surrogate gradient computations are pre-compiled into machine-level codelets, and concurrently processed with the neural network (NN) matrix-vector-multiplication step.}
    \label{fig:3}
\end{figure*}

%%%%%%% Include a figure?
%%%%%%% Include pseudo-code?
The `Poplar SDK' interfaces popular deep learning frameworks directly into IPU programming. The IPU uses an autodifferentiation engine independently of PyTorch's backend, and as such, spiking neuron models that depend on surrogate gradient descent are not compilable by default. Custom operations must be written in C++ and pre-compiled into machine-level codelets that are accessible to users via Python. 

Our approach pre-compiles the surrogate gradient operator at the time \textit{snnTorch} is imported. A custom operation is defined for the threshold-shifted Heaviside function (see (3)) implemented in C++, which is compiled thus generating a shared library object that can be dynamically linked in Python at runtime. This allows for the IPU-build of \textit{snnTorch} to be syntactically near identical to CPU/CUDA-based usage, abstracting away machine-level complexities from the user. The surrogate gradient operator is co-located in the same IPU core which reduces the impact of non-modular function calls that are needed when overriding the autograd module in PyTorch. This is sequenced via pseudo-code in Algorithm 1 and illustrated in \Cref{fig:3}.

% We take advantage of this feature by developing our own pre-compiled surrogate gradient operators that accelerate components of the backward pass that are not modular, and pose the risk of being the critical path during training. % (detailed in \Cref{sec:custom}).

% Graphcore’s SDK Poplar has been designed to allow for PyTorch to be implemented and used on the IPU for optimal performance. There is also the option to implement custom operations that can be used in Python. 

Specifically, (3) is a non-differentiable function. This function is replaced in the backward pass with the user's choice of approximation. For example, a straight-through-estimator simply bypasses the non-differentiable operator \cite{hinton2012neural}. Alternative approaches use functional approximations of the Heaviside operator by smoothing out the discontinuous step, e.g., the fast-sigmoid function:

\begin{equation} \label{eq:fs}
    \tilde{z} = \frac{(u-u_{\rm thr})}{(1+|u-u_{\rm thr}|)},
\end{equation}
\begin{equation}\label{eq:dfs}
    \frac{\partial z}{\partial u}\leftarrow\frac{\partial \tilde{z}}{\partial u}=\frac{1}{(1+|u_{\rm thr} - u|)^2},
\end{equation}

\noindent where the left-arrow denotes substitution, and the tilde in $\tilde{z}$ represents an approximation.

\subsection{Network Architecture}

%\begin{figure}[!t]
    %\centering
    %\includegraphics[width=0.5\textwidth]{Figures/Picture2.png}
    %\caption{The network architectures tested for this paper. The top image is the %Fully-Connected Network while the bottom image is the Convolutional Neural Network.}
    %\label{fig:throughput1}
%\end{figure}

For this paper, two network types were tested on four different types of hardware. The hardware tested include: the NVIDIA A100, NVIDIA V100, NVIDIA GTX 1080, and the Graphcore IPU Mk2. The networks tested are designed to fit a single processor to avoid comparisons that are I/O-limited. The architectures include: a 3-layer dense SNN (DSNN) and a 3-layer convolutional SNN (CSNN). Despite the small size of the networks, these were trained over multiple time steps which led to near-full memory utilization. Leaky integrate-and-fire neurons are used for all experiments unless otherwise specified, and most spiking simulations are performed across 25 time steps. For experiments measuring throughput, the MNIST dataset is used in the interest of speed \cite{lecun1998mnist}. For experiments that account for loss-based metrics (e.g., accuracy), CIFAR-10 is used \cite{krizhevsky2009learning}.
The various architectures used are specified in \Cref{tab:arch}. 5C12 refers to a $5\times5$ convolutional kernel with 12 channels. MP2 refers to a $2\times2$ max-pooling operator. Unless otherwise specified (e.g., in experiments that sweep across different architecture parameters), these networks are used for the experiments that follow with the AdamW optimizer used in all cases \cite{kingma2014adam}. Where relevant, experiments were repeated five times to generate error bars.

%% Add footnote to describe meaning of 5C12 and MP2
\begin{table}[!t]
    \centering
    \caption{Network Architecture}
    \begin{tabular}{l|c}
        \textbf{Network} & \textbf{Architecture}\\
        \hline
        DSNN & 784--1000--10\\
        \hline
        CSNN & 5C12--MP2--5C64--MP2--10\\
        \hline
    \end{tabular}
    \label{tab:arch}
\end{table}

% SLSTM? 

\section{Experimental Results} \label{sec:exp}
The following experiments have been conducted to benchmark IPU performance:
\begin{itemize}
    \item Baseline FLOPS (floating point operations per second)
    \item Baseline Throughput
    \item Throughput across batch sizes
    \item Throughput across architectures
    \item Throughput across neuron models
    \item Compute time spent on spiking vs. static dynamics
    \item Mixed precision throughput
    \item Population coding: throughput and accuracy
    \item Power usage per operation
\end{itemize}

All experiments that follow account for the entire training process using BPTT, including the forward-pass, gradient calculation, and weight update.

\subsection{Baseline FLOPS}
Before performing IPU vs. GPU performance comparisons, we first assess the performance of a spiking network against equivalent, non-spiking artificial neural networks (ANNs) on the IPU.% during inference. 
One FLOP is defined as one fused multiply-add floating point operation, calculated using the \textit{fvcore} Python Library. The FLOPs comparison can be seen in \Cref{tab:flops}. On average, the IPU improves TFLOPS by $4.6\times$ when compared to the A100, $6.4\times$ over the V100, and $10\times$ over the GTX1080. %For the non-spiking case, while the GPUs remain the same. 

Interestingly, the performance of the spiking network is marginally better for the dense case than the non-spiking network. This may be because the IPU has been optimized for handling different types of concurrent operations, where processing neuron state-based computations are relatively simple operators when compared to large-scale matrix-vector multiplication. On the other hand, the TFLOPS when running the convolutional SNN drops by approximately 57\% from non-spiking to spiking networks on the IPUs.

%TABLE I. 	FLOP COMPARISON BETWEEN SNN AND ANN
%Network Type	FLOPs
%Two-layer SNN	2.541G
%Two-layer ANN	0.102G
%Three-layer SNN	8.134G
%Three-layer CNN	0.325G
% \begin{table}[h!]
%     \centering
%     \caption{SNN vs ANN TFLOPS}
%     \begin{tabular}{l|c|c|c}
%         \textbf{Network Type} & \textbf{IPU} & \textbf{GTX1080} & \textbf{V100}\\
%         \hline
%         DNN & 1.02 & 0.10 & 0.16 \\
%         \textbf{DSNN} & \textbf{25.97} & \textbf{2.60} & \textbf{4.04}\\
%         \hline
%         CNN & 3.24 & 0.33 & 0.51 \\
%         \textbf{CSNN} & \textbf{46.37} & \textbf{4.65} & \textbf{7.22}\\
%         \hline
%     \end{tabular}
%     \label{tab:flops}
% \end{table}

\begin{table}[h!]
    \centering
    \caption{SNN vs ANN TFLOPS (Training)}
    \begin{tabular}{l|c|c|c|c}
        \textbf{Network Type} & \textbf{IPU} & \textbf{A100} & \textbf{V100} & \textbf{GTX1080}\\
        \hline
        DNN & 1.02 & 0.22 & 0.16 & 0.10  \\
        DSNN & 1.04 & 0.22 & 0.16 & 0.10 \\
        \hline
        CNN & 3.24 & 0.70 & 0.51 & 0.33 \\
        CSNN & 1.85 & 0.40 & 0.29 & 0.19\\
        \hline
    \end{tabular}
    \label{tab:flops}
\end{table}

\subsection{Baseline Throughput}
%%%%%% Fig 5, 7,8, 12 needs to be divided up into DSNN / CSNN
Throughput is measured in terms of 1000s of images per second, and accounts for the wallclock time commencing from the forward pass, the backward pass, and concludes once the weight update is completed. A batch size of 128 images is used by default.  Each network is trained over 60 epochs. To obtain error bars, this is repeated 20 times on each hardware. The throughput is calculated by:
\begin{itemize}
  \item measuring the wallclock time to process one minibatch,
  \item dividing the batch size by the wallclock time.
\end{itemize}

\subsubsection{DSNN Throughput} 
The results from the DSNN are tabulated in \Cref{tab:dsnn_tp}. The IPU can train an average of 46,297 images per second, which is 3.1$\times$ higher than the A100, 6.4$\times$ higher than the V100, and 9.9$\times$ higher than the GTX1080. %, and 79$\times$ greater than the CPU. 
Error bars across multiple trials are illustrated in \Cref{fig:thruput}(a). The standard deviation for the IPU is approximately 3,623 images.% , which maintains a large margin of improvement.
% Once the higher power draw of the IPU is accounted for, the improvement in throughput/watt is approximately 4--5$\times$ over the GPU alternatives.

\subsubsection{CSNN Throughput}
With respect to the CSNN, there is a much larger number of computations being performed leading to a decrease in throughput for both IPUs and GPUs. The IPU training throughput is 15,566 images per second. This is 2.1$\times$ more than the A100, 4$\times$ higher than the V100, and 5.9$\times$ greater than the GTX108. % 128$\times$ higher than the CPU. 
The standard deviation is 1,069 images per second.

The performance margin between the DSNN and CSNN indicates that the IPU has been optimized for high memory usage. 
This is useful for experiments that require traces of membrane potential to be stored as with BPTT, for pre- and post-synaptic current traces as with spike time-dependent plasticity \cite{bi1998synaptic}, and also for dynamically varying synapses.

\begin{table}[!t]
    \centering
    \caption{DSNN Throughput}
    \begin{tabular}{l|c|c|c|c}
        \textbf{Metric} & \textbf{IPU}  & \textbf{A100} & \textbf{V100} & \textbf{GTX1080} \\
        \hline
        Average Power & 92.12W & 60.51W & 55.71W & \textbf{49.29W} \\
        \hline
        Average Throughput & \textbf{46297.17} & 9858.92 & 7207.24 & 4639.89 \\
        (images/s)         &                   &                   & \\
        \hline
        Throughput/Watt & \textbf{502.59} & 162.93 & 129.4 & 94.4 \\
        \hline
    \end{tabular}
    \label{tab:dsnn_tp}
\end{table}

\begin{table}[!t]
    \centering
    \caption{CSNN Throughput}
    \begin{tabular}{l|c|c|c|c}
        \textbf{Metric} & \textbf{IPU} & \textbf{A100} & \textbf{V100} & \textbf{GTX1080} \\
        \hline
        Average Power Used & 92.57W  & 68.86W & 61.99W & \textbf{55.44W}\\
        \hline
        Average Throughput & \textbf{15566.16} & 5608.43 & 3883.89 & 2635.54\\
        (images/s)         &                   &         & \\
        \hline
        Throughput/Watt & \textbf{168.16} & 81.44 & 62.65  & 47.54 \\ 
        \hline
    \end{tabular}
    \label{tab:csnn_tp}
\end{table}

%%%% up to here, about to start explaining improvement over batch sizes
\subsection{Throughput Across Batch Size}
\begin{figure}[!t] 
    \centering
  \subfloat[]{%
      \includegraphics[width=0.5\linewidth]{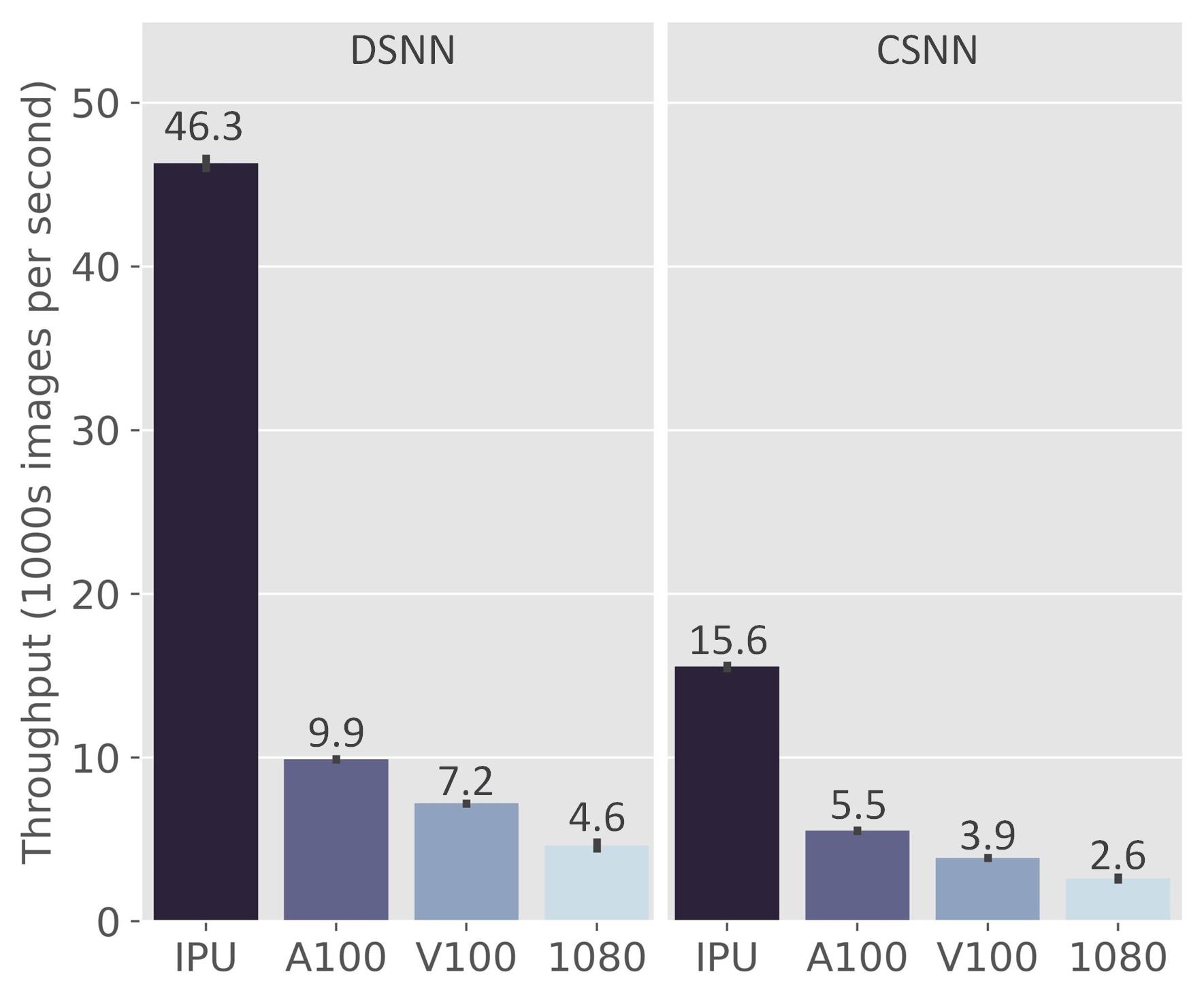}}
    \hfill
  \subfloat[]{%
        \includegraphics[width=0.5\linewidth]{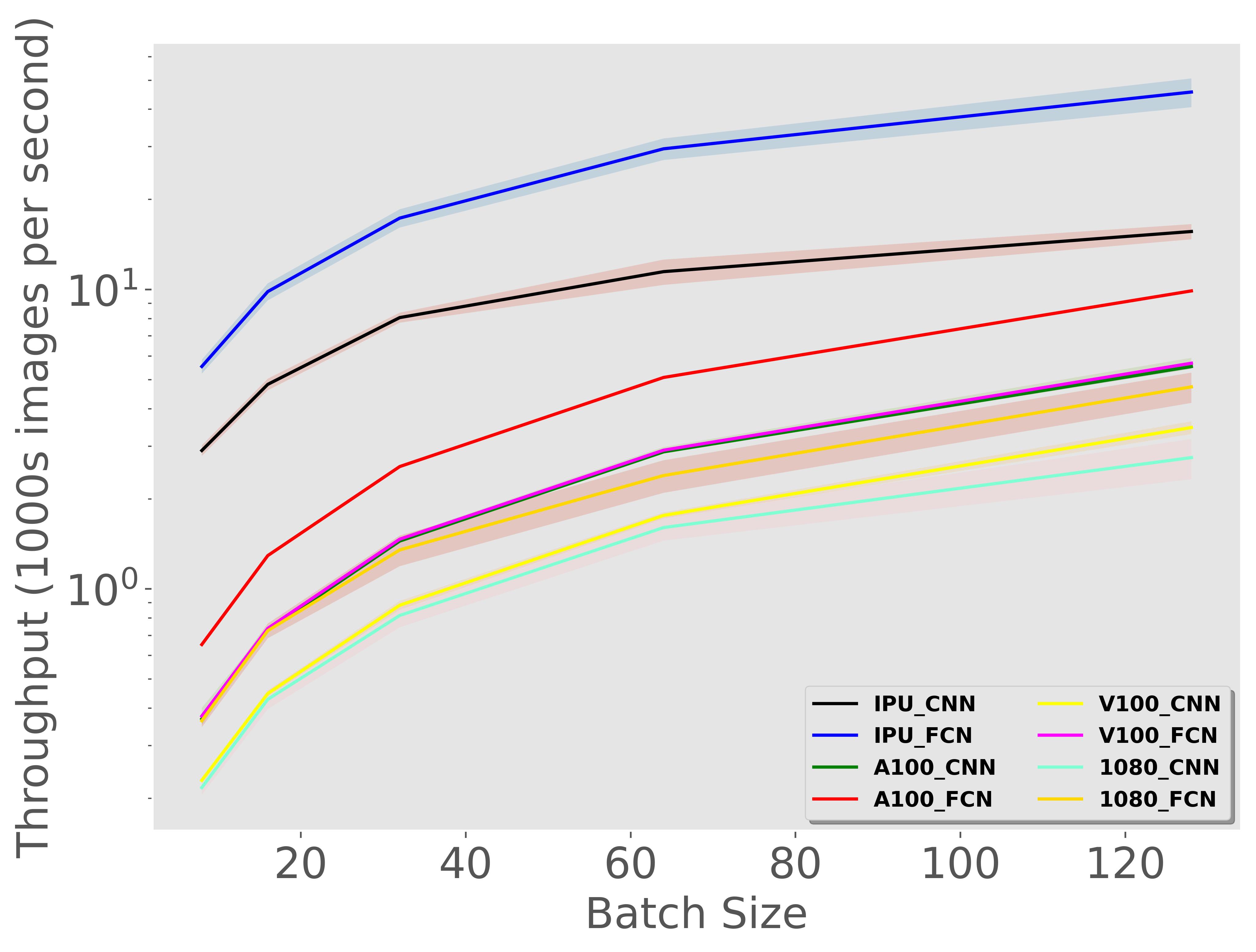}}
%     \\
%   \subfloat[c\label{1c}]{%
%         \includegraphics[width=0.45\linewidth]{example-image}}
%     \hfill
%   \subfloat[d\label{1d}]{%
%         \includegraphics[width=0.45\linewidth]{example-image}}
  \caption{(a) Baseline throughput. (b) Throughput with varying batch size.}
  \label{fig:thruput} 
\end{figure}

As networks increase in size, memory limits constrain the maximum possible batch size that is permissible. This problem is exacerbated in SNNs which also consume memory for each additional simulated time step. To measure this effect, the batch size was swept from 8 to 128, with throughput results shown in \Cref{fig:thruput}(b). On inspection, there is far less variance in performance for IPUs. This is especially important where a large number of time steps must be simulated, and the maximum batch size decreases. 
Close attention is given to the smallest tested batch size, as real-world batch sizes in continual learning workloads are `1'. 
For the smallest tested batch size ($n=8$), the performance improvement of the IPU over the A100 for both CSNN and DSNN is more than one order of magnitude (14$\times$). 

% With different batch sizes, the IPU again outperforms the GPU, as seen in Figure 3. For a FCN, it can be seen that the IPU can achieve 14 times to 9 times higher throughput than the GPUs with batch sizes from 8 to 128, respectively, while for a CNN, the IPU can achieve a speed up of 14 to 5.7 times compared to the GPUs with batch size varying from 8 to 128.

%%% This will provide a baseline for how the different hardware processes training data. To establish our baseline, a batch size of 128 is used to measure training throughput.

\begin{figure} 
    \centering
  \subfloat[]{%
      \includegraphics[width=0.5\linewidth]{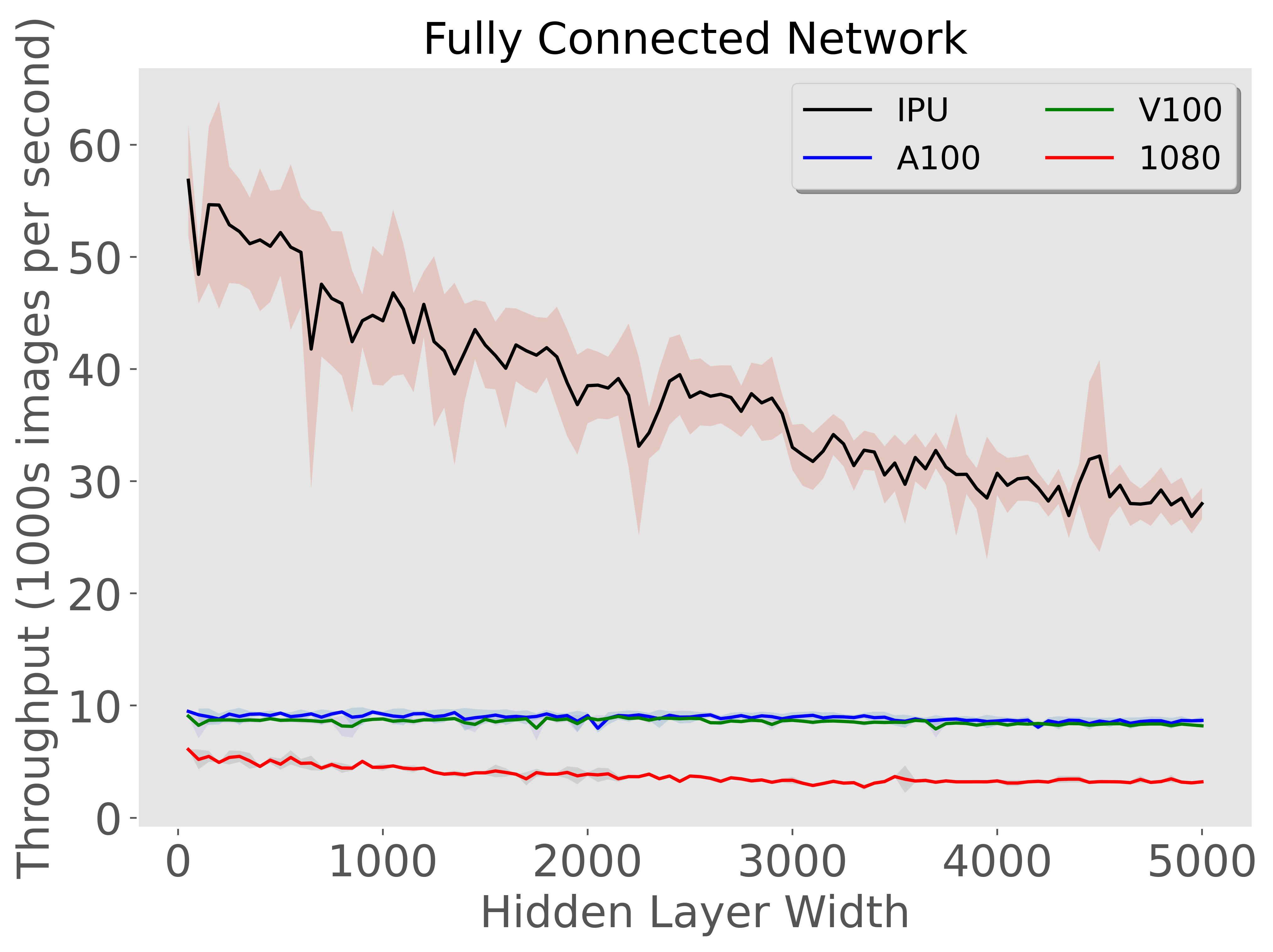}}
    \hfill
  \subfloat[]{%
        \includegraphics[width=0.5\linewidth]{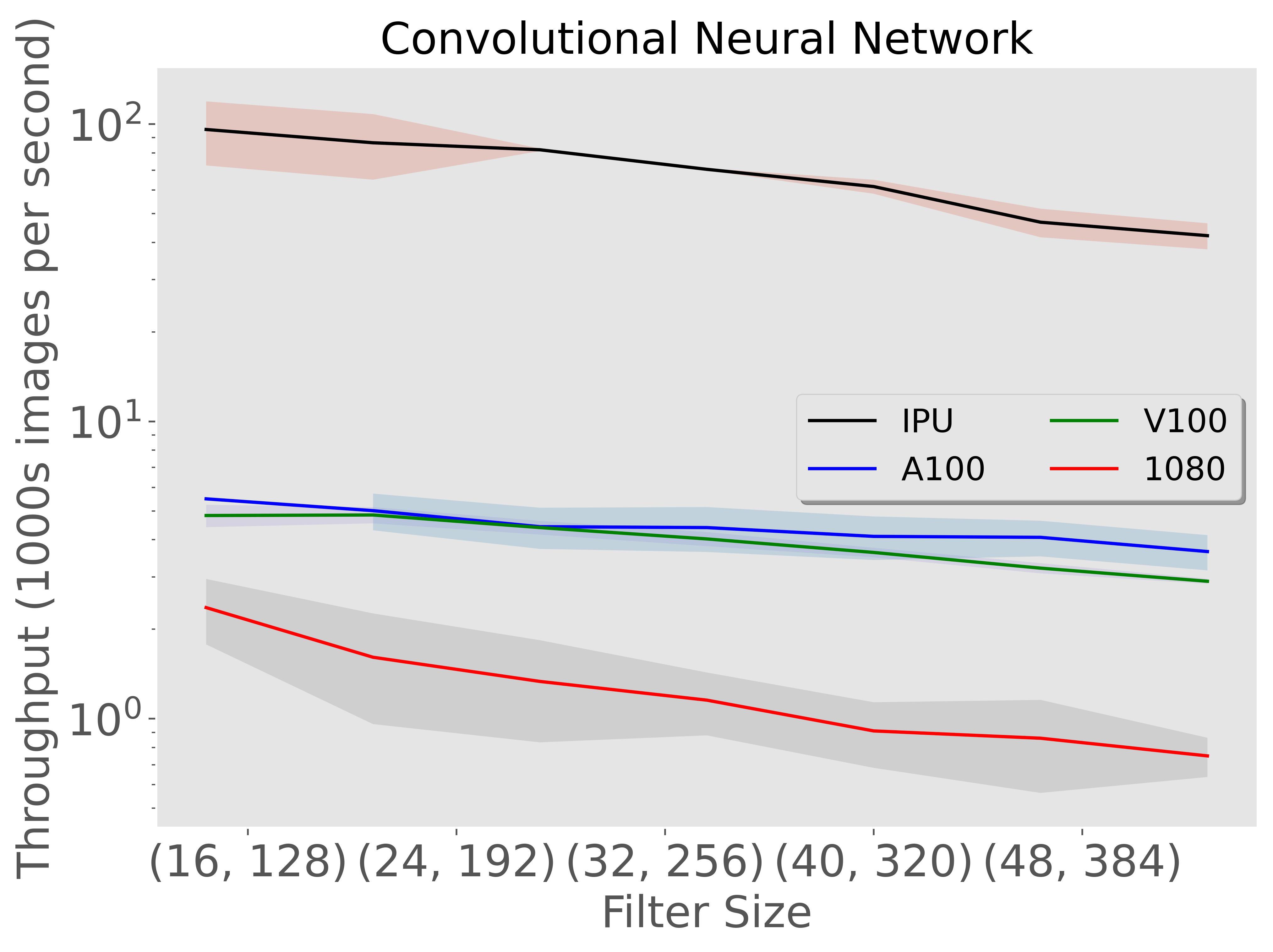}}
%     \\
%   \subfloat[c\label{1c}]{%
%         \includegraphics[width=0.45\linewidth]{example-image}}
%     \hfill
%   \subfloat[d\label{1d}]{%
%         \includegraphics[width=0.45\linewidth]{example-image}}
  \caption{Throughput with varying network architectures. (a) DSNN with increasing network width. (b) CSNN with increasing convolutional kernel depth. For ($N_1$, $N_2$), $N_1$ corresponds to depth of the first layer kernel, and $N_2$ is the depth of the second layer kernel.}
  \label{fig:arch} 
\end{figure}

\subsection{Throughput Across Architectures}
Network architecture is varied for both the DSNN and CSNN and throughput is measured. For the DSNN, the number of neurons in the hidden layer is increased, and for the CSNN, the kernel depths of the first two convolutional filters are increased.

\subsubsection{DSNN Throughput} 
GPUs are completely insensitive to increasing the number of neurons, as shown in \Cref{fig:arch}(a). This indicates that for a small network, a large number of cores available are underutilized. On the other hand, the margin of improvement with the IPU increases with smaller networks. This is because different operations can be parallelized to improve utilization of the large number of IPU cores available.

\subsubsection{CSNN Throughput}
The throughput of varying CSNN architectures is illustrated in \Cref{fig:arch}(b). In contrast to DSNNs, larger convolutional kernels decrease the throughput of GPUs. The larger number of computations involved in convolutions indicates that the GPU cores are now fully utilized.

\begin{figure}[!t]
    \centering
  \subfloat[]{%
      \includegraphics[width=0.5\linewidth]{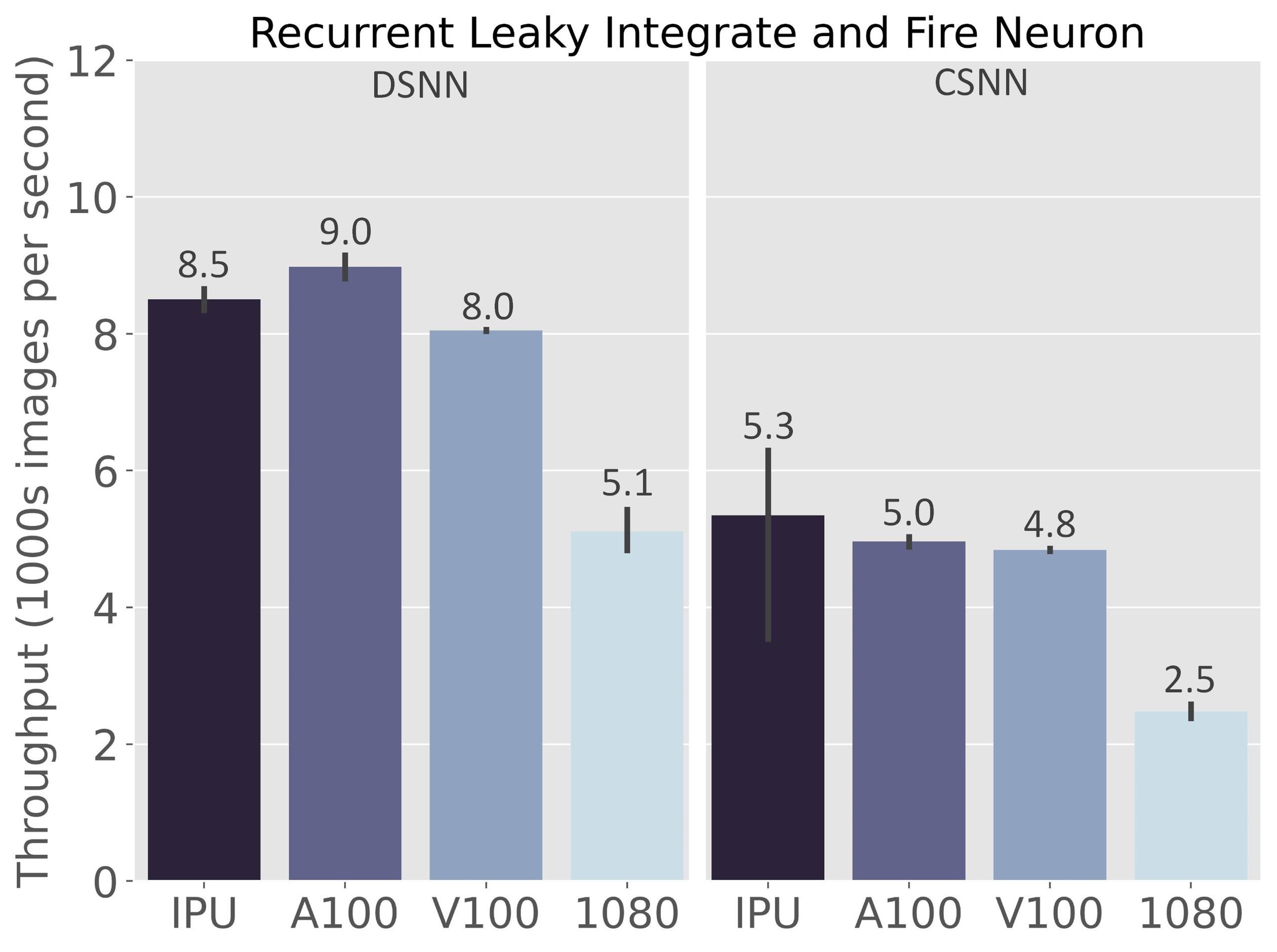}}
    \hfill
  \subfloat[]{%
        \includegraphics[width=0.5\linewidth]{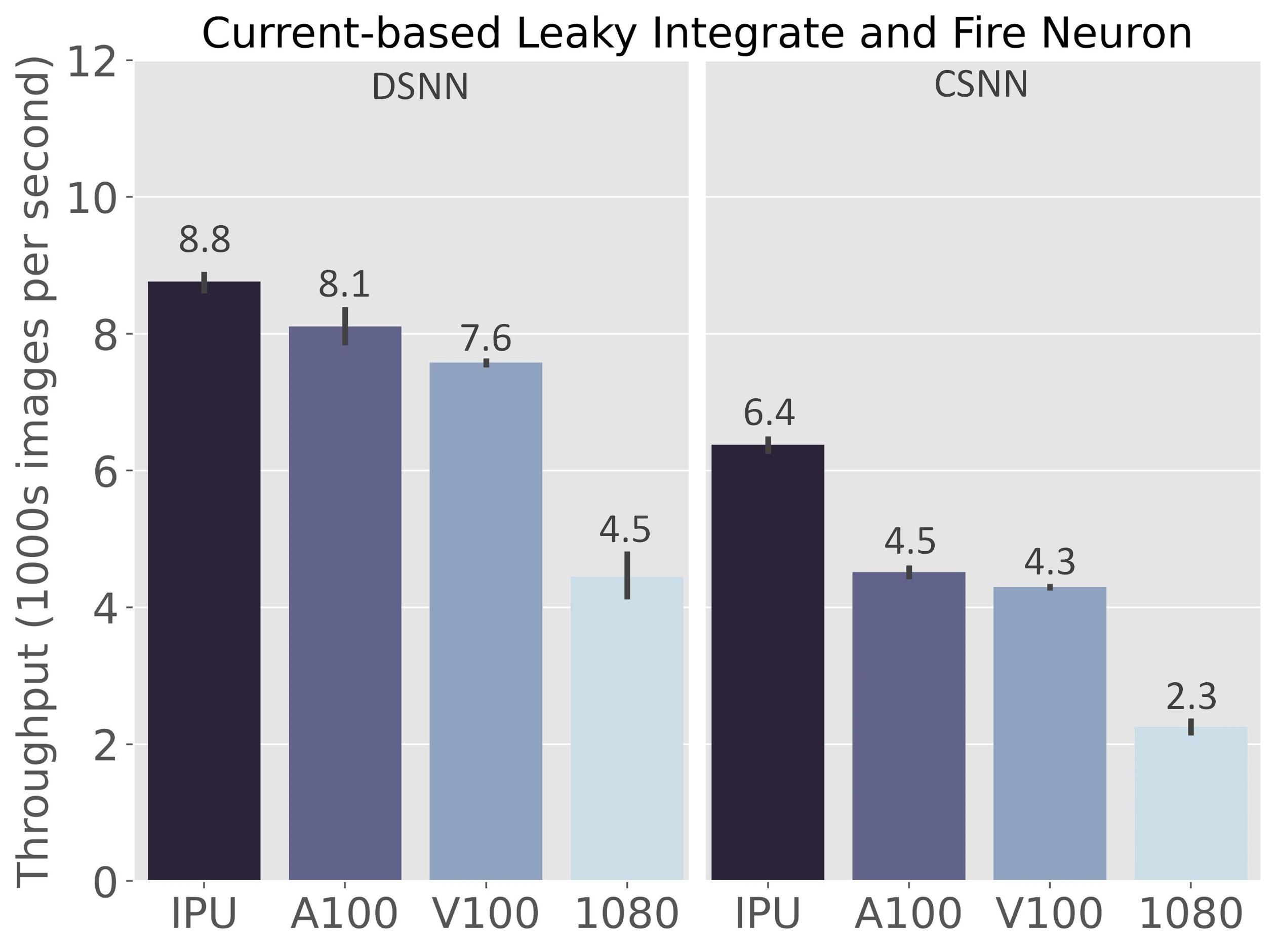}}
%     \\
%   \subfloat[c\label{1c}]{%
%         \includegraphics[width=0.45\linewidth]{example-image}}
%     \hfill
%   \subfloat[d\label{1d}]{%
%         \includegraphics[width=0.45\linewidth]{example-image}}
  \caption{Throughput with alternative neuron models. (a) Recurrent SNN. (b) Current-based Neuron Model.}
  \label{fig:neurons} 
\end{figure}

\subsection{Alternative Neuron Models}
Several other spiking neuron models are increasing in usage in the context of SNNs. Recurrent spiking neuron models, e.g., (7), have been shown to achieve better performance on datasets with temporal complexity \cite{perez2021neural}. Current-based neuron models, e.g., see (5) and (6), are better suited for learning precise spike timing, as the membrane potential trace is differentiable with respect to time. Throughput for a recurrent SNN is shown in \Cref{fig:neurons}(a), and that of an SNN composed of current-based neurons is shown in \Cref{fig:neurons}.

The performance of V100s remains relatively unaffected by more complex neuron models, which causes the performance gap with IPUs to narrow. This highlights a potential opportunity to improve resource allocation during compilation. There are more steps to process these more exotic neuron models, and so more cores are allocated to handling those operations. This comes at the cost of less resources available to process synaptic operations, where computational complexity scales with $\mathcal{O}(n^2)$.

% \begin{figure}[!t]
%     \centering
%     \includegraphics[width=0.5\textwidth]{Figures/width_fcn.jpg}
%     \caption{FCN with varying hidden neurons}
%     \label{fig:throughput7}
% \end{figure}

% \begin{figure}[!t]
%     \centering
%     \includegraphics[width=0.5\textwidth]{Figures/width_cnn.jpg}
%     \caption{CNN with varying hidden neurons}
%     \label{fig:throughput8}
% \end{figure}

%%%%%% continue from here
\subsection{Spiking vs. Static Dynamics}
\begin{figure}[!t]
    \centering
  \subfloat[]{%
      \includegraphics[width=0.47\linewidth]{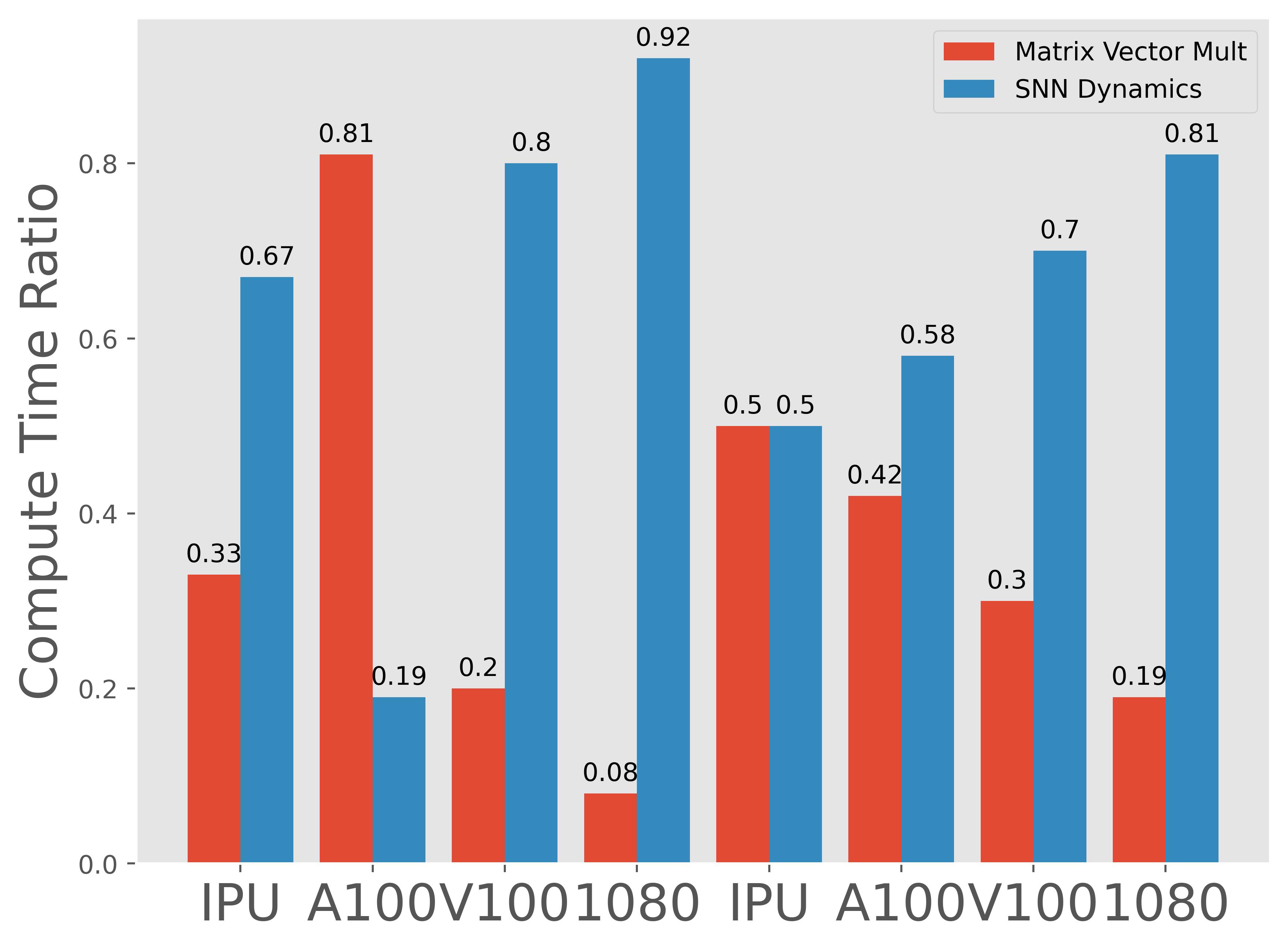}}
    \hfill
  \subfloat[]{%
        \includegraphics[width=0.53\linewidth]{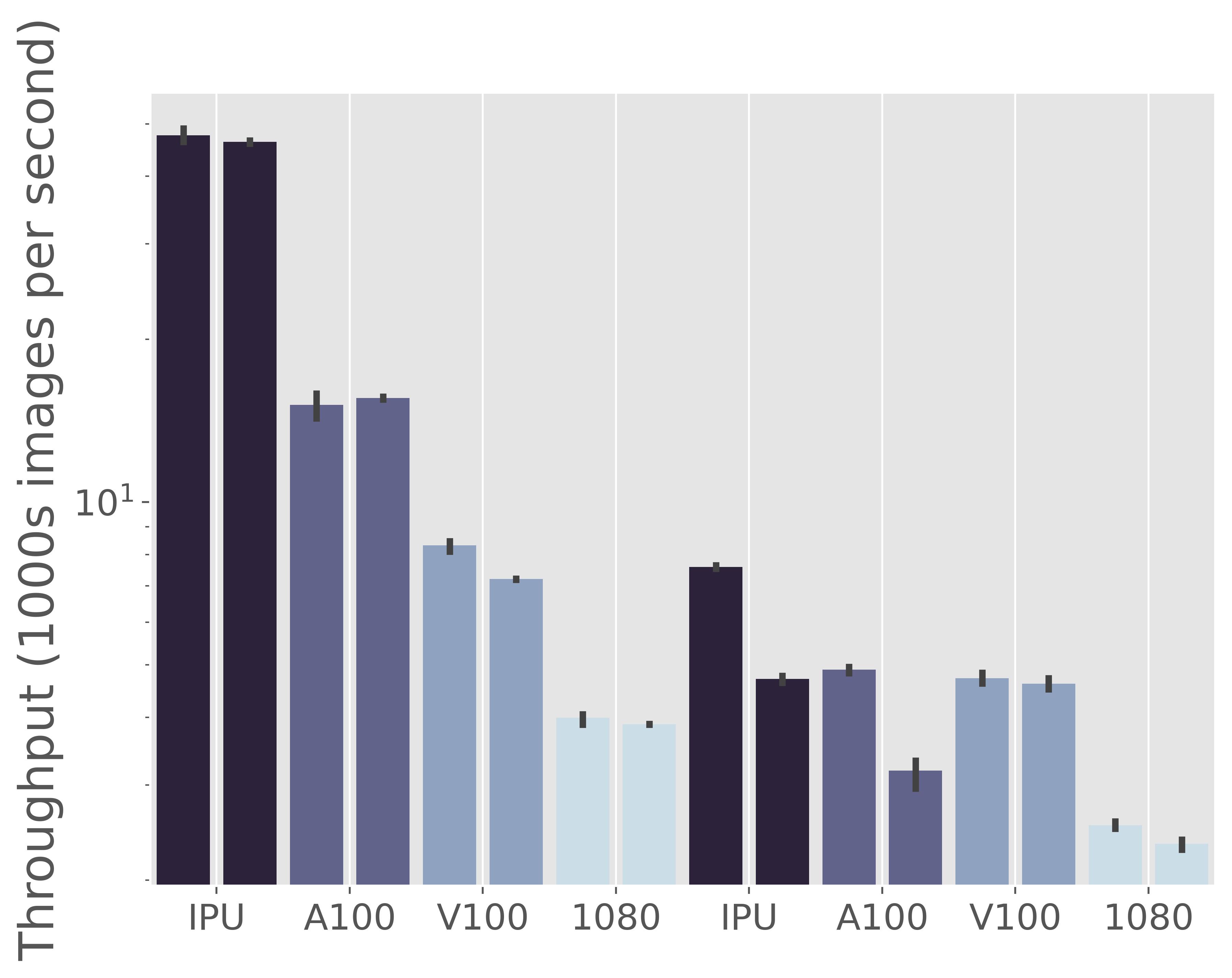}}
%     \\
%   \subfloat[c\label{1c}]{%
%         \includegraphics[width=0.45\linewidth]{example-image}}
%     \hfill
%   \subfloat[d\label{1d}]{%
%         \includegraphics[width=0.45\linewidth]{example-image}}
  \caption{(a) Compute time ratio of spiking dynamics (red) and synaptic opertions/matrix-vector multiplications (blue). (b) Throughput with full precision (32-bit: right) and half precision (16-bit: left).}
  \label{fig:ratio} 
\end{figure}

To verify the above theory, the ratio of time spent calculating the dynamics of spiking neurons (i.e., solving (3) and (4)) is compared against the amount of time spent on matrix-vector multiplication. The results are shown in \Cref{fig:ratio}(a), demonstrating that IPUs provide better balance between neuronal and synaptic operations. In the CSNN, the amount of compute time allocated to solving state-driven dynamics is exactly equivalent to the duration of time spent on synaptic operations. This is beneficial for simple neuron models, but where more complex neurons are concerned, may require further optimization during compile time. Further improvements could be obtained by exploiting function outlining which merges repeatable code-blocks for execution on identical cores in IPUs. This can reduce the overhead allocated to solving state dynamics, and free up more cores to run synaptic operations.

% For the ratio of compute time spent on the network, it can be seen from Figure 9 the IPU has a more balanced spread between computing the traditional NN layers and the spiking neurons, with the CNN achieving almost 50:50 split, compared to the GPUs where most of the computation is spent on the spiking dynamics.

\subsection{Mixed Precision Performance}
Mixed precision training reduces the bit-width needed for computation, which comes with an associated wallclock time reduction. This is often with minimal, if any, impact on network performance. The default full precision (32b) mode is compared to half precision (16b) training, with results shown in \Cref{fig:ratio}(b). The difference for all cases is marginal because gradients continue to be calculated in full precision.

% \begin{figure}[!t]
%     \centering
%     \includegraphics[width=0.5\textwidth]{Figures/Time_Boxplot.jpg}
%     \caption{Compute Time\%}
%     \label{fig:throughput9}
% \end{figure}

%  \begin{figure}[!t]
%     \centering
%     \includegraphics[width=0.5\textwidth]{Figures/Float_Boxplot.jpg}
%     \caption{Throughput with 16-bit FP and 32-bit FP}
%     \label{fig:throughput10}
% \end{figure}
 
 \begin{figure}[!t]
    \centering
  \subfloat[]{%
      \includegraphics[width=0.5\linewidth]{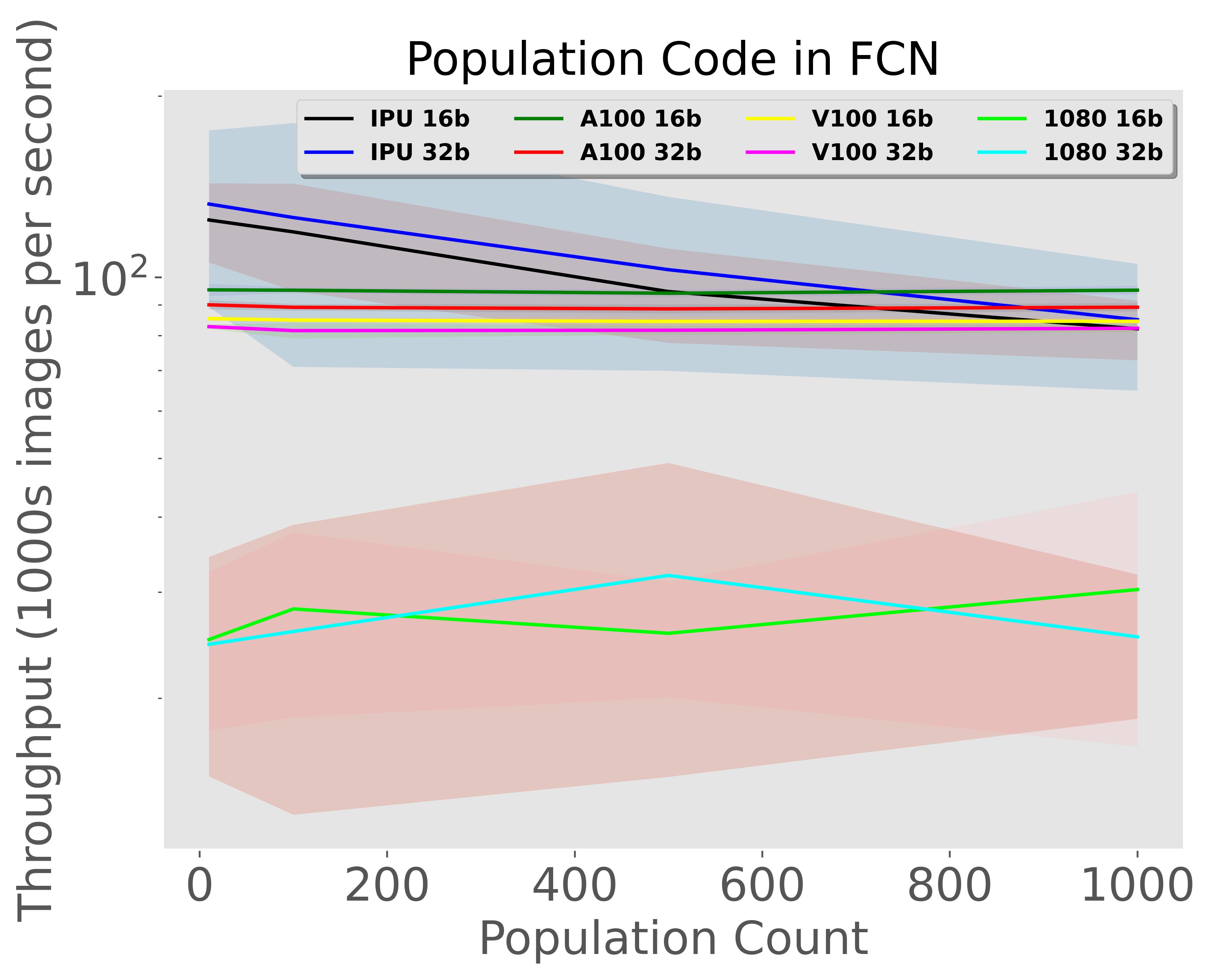}}
    \hfill
  \subfloat[]{%
        \includegraphics[width=0.5\linewidth]{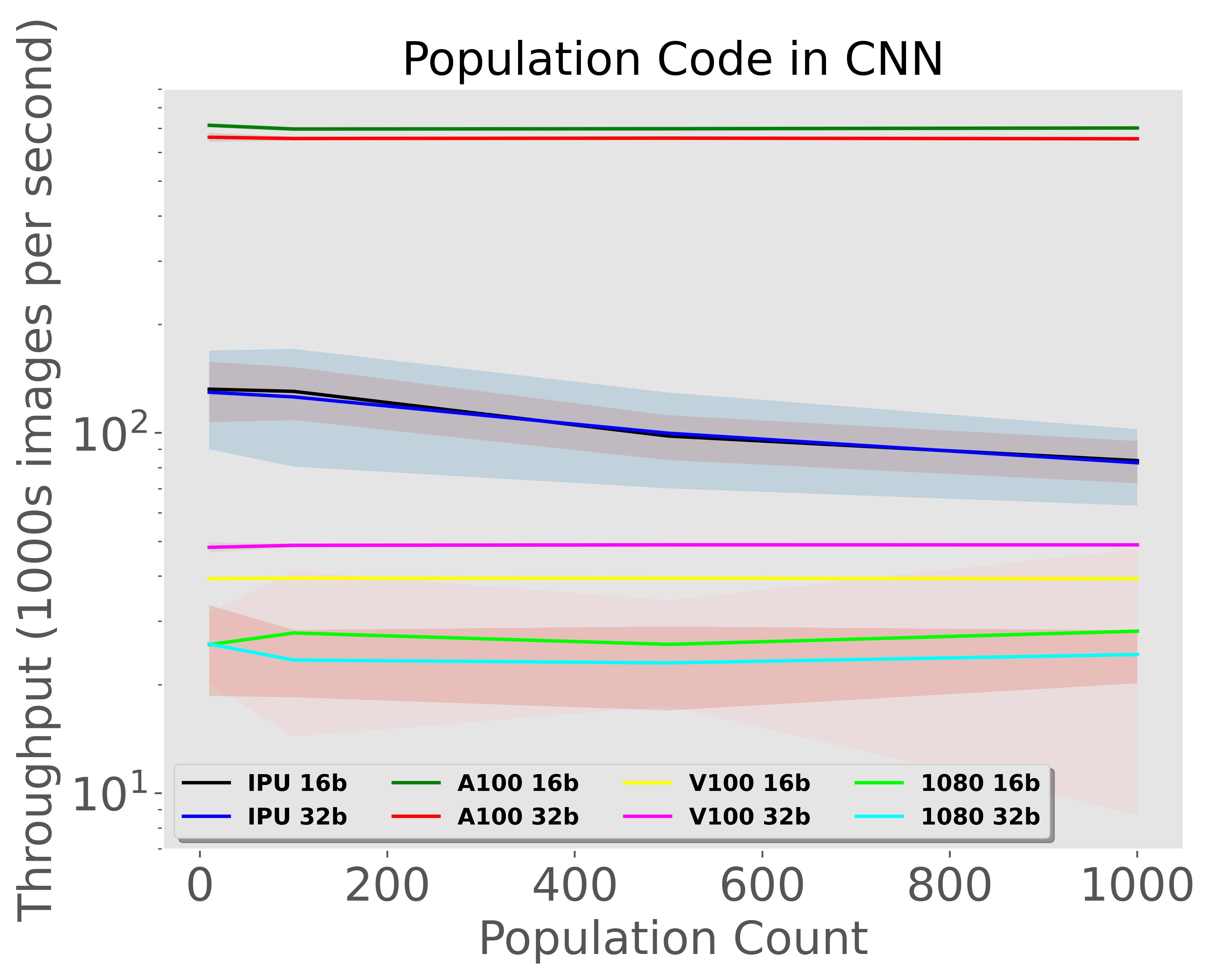}}
%     \\
%   \subfloat[c\label{1c}]{%
%         \includegraphics[width=0.45\linewidth]{example-image}}
%     \hfill
%   \subfloat[d\label{1d}]{%
%         \includegraphics[width=0.45\linewidth]{example-image}}
  \caption{Population code throughput. (a) DSNN with population coding. (b) CSNN with population coding.}
  \label{fig:pop} 
\end{figure}

%% Make this its own section?
\section{Acceleration Using Population Codes}
\subsection{Biological Plausibility}
At present, the most common approach to determining the predicted class is to select the neuron with the highest firing count. This is equivalent to using a rate coded SNN. In neurophysiology, it is thought that rate codes alone cannot be the dominant encoding mechanism in the primary cortex. One of several reasons is because the background neuronal firing rate is roughly 0.1 -- 1~Hz, which is far slower than the reaction response time of animals and humans.

But if multiple neurons are grouped with their collective spikes counted cumulatively, then it becomes possible to measure a firing rate for a population of neurons in a very short window of time. Assigning a population of neurons to individual classes is also known as using a `population code'. Population coding adds credibility to the biological plausibility of rate-encoding mechanisms.

\subsection{Population Codes in Unsupervised Learning}
In the past, it has been common practice to increase the number of neurons at the output layer of a network, and cluster the response of various neurons together.
This practice has been limited to networks trained using spike timing-dependent plasticity.
Neurons would be assigned classes based on which assignments led to the highest accuracy.
As such, using a population code where multiple neurons were assigned per class typically led to a boost in classification accuracy in unsupervised learning tasks. This is because more neurons means more permutations of neuron assignments that can increase accuracy. 
The shift from unsupervised learning to gradient-based supervised learning has made population codes a diminishing practice when training SNNs, as targets are pre-assigned before training commences. We find that using population codes offers alternative benefits when training SNNs.

\subsection{Population Codes in Gradient-based Learning}
These benefits are grounded in the fact that accelerators are optimized for parallel operations rather than sequential operations. Using a population of neurons redistributes the time cost over space instead, i.e., larger dimension matrix-vector multiplications can be used instead of repeatedly applying matrix-vector multiplications with smaller dimensions. We run a series of experiments to show population codes further accelerate throughput with a marginal impact on accuracy. Because accuracy is now of interest, we assess performance on the CIFAR-10 dataset as MNIST is broadly recognized as being too simple.

\subsubsection{Experimental Setup}
Similar network architectures as described in \Cref{tab:arch} are used. For the DSNN, the number of input neurons is increased due to the larger dimensionality of the CIFAR-10 dataset (32$\times$32$\times$3) over that of the MNIST dataset (28 $\times$ 28 $\times$ 1). The same holds true for the terminal layer of the CSNN.

\subsubsection{Training Throughput}
A comparison of throughput across various output neurons and with different precision (half and full) are shown in \Cref{fig:pop}. One single simulation time step is used. Performance follows a very similar trend to that of varying network architectures in \Cref{fig:arch}, where GPUs perform identically as the output population increases. At the IPU's best, optimal throughput is approximately 145,000 images per second. This is 37$\times$ better than the original baseline performance (despite using bigger images with 3 channels), and approximately twice as fast as the best GPU. 
At its lowest throughput, performance of the IPU and A100 converges in the DSNN experiment. When population codes are applied to CSNNs, the A100 skyrockets in performance and becomes invariant to architectural changes. The large number of terminal synaptic operations dominates the total cost of the network, completely outweighing state-based neuronal operations. This places the A100 in the lead in population-based CSNN benchmarks.

\subsubsection{Accuracy}
\begin{figure}[!t]
    \centering
    \includegraphics[width=0.35\textwidth]{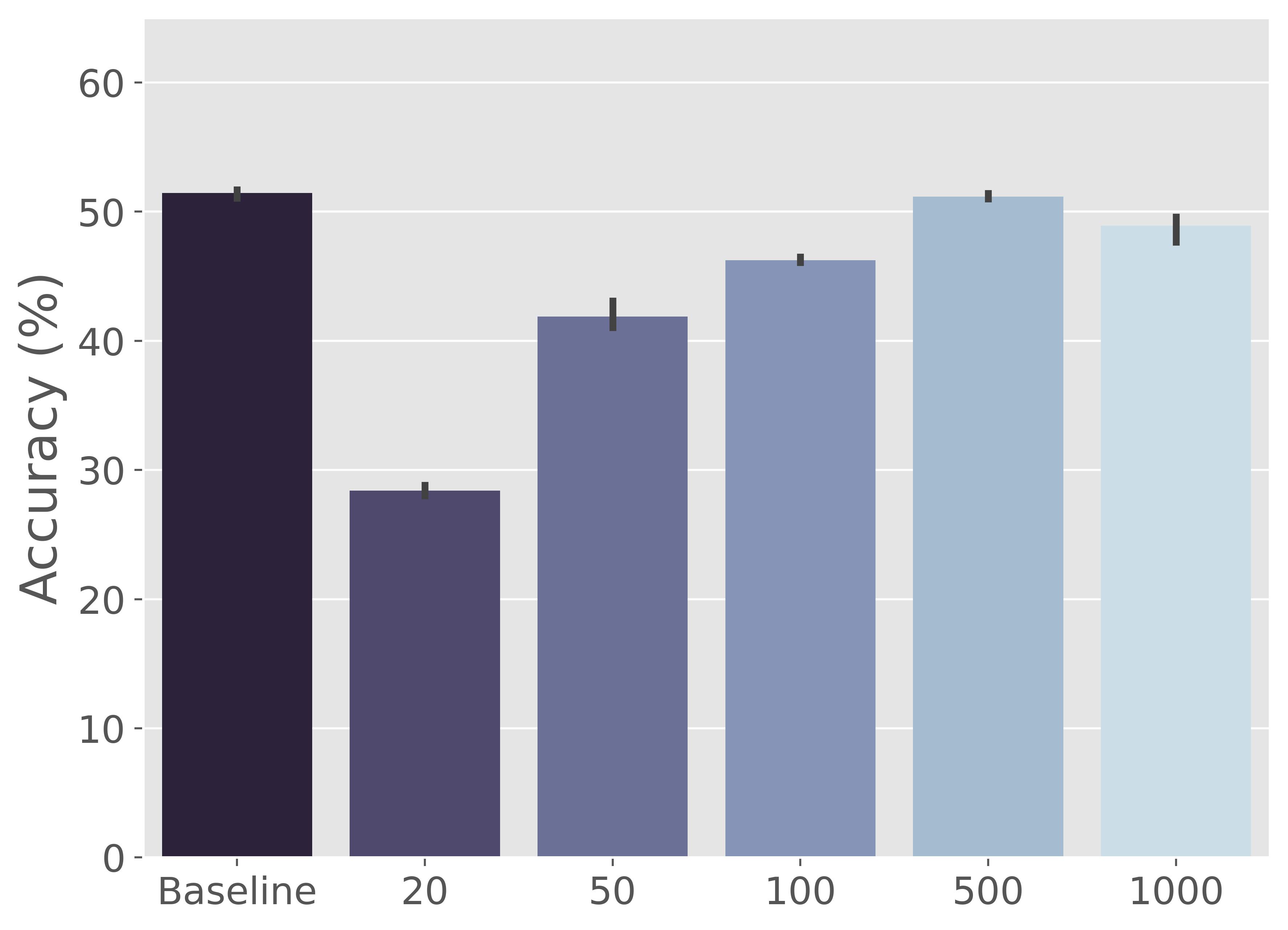}
    \caption{Accuracy Performance with Population Coding.}
    \label{fig:pop_acc}
\end{figure}

As a coarse-grain measure of accuracy, the DSNN model was used to provide an idea as to how population codes impact training performance. The DSNN is trained over 5 epochs to determine whether it is possible to train networks in one single time-step, where each neuron is constrained to only firing a maximum of once. Results are illustrated in \Cref{fig:pop_acc}, where a baseline accuracy of 52.2\% is obtained without using population codes (i.e., 10 output neurons simulated over 25 time steps). This accuracy is almost reached when 500 output neurons are used, assigning 50 output neurons per class. As a matter of interest, indefinitely increasing the output neuron count does not continue to increase performance. Based on prior approaches to constructing models, network depth should be increased commensurately to network width to avoid leaning towards either end of the bias-variance trade-off \cite{zagoruyko2016wide}.

We note that the target here is not state-of-the-art accuracy, but rather, to assess whether single time-step learning is possible at all. Our results indicate that equal performance to multiple time-steps can be met by using population codes, verified on a simple DSNN architecture.

% The performance boost from population coding may start to fade as the number of time steps increases. 

%  \begin{figure}[!t]
%     \centering
%     \includegraphics[width=0.5\textwidth]{Figures/population_fcn.jpg}
%     \caption{FCN with Population Coding}
%     \label{fig:throughput11}
% \end{figure}
 
% \begin{figure}[!t]
%     \centering
%     \includegraphics[width=0.5\textwidth]{Figures/population_cnn.jpg}
%     \caption{CNN with Population Coding}
%     \label{fig:throughput12}
% \end{figure}

%\begin{figure}[!t]
    %\centering
    %\includegraphics[width=0.5\textwidth]{Figures/Sparsity.jpg}
    %\caption{Throughput of Different Input Sparsity}
    %\label{fig:throughput13}
%\end{figure}
 
% % We trained the two different SNNs with three different hardware, Nvidia GTX 1080, Nvidia V100 and Graphcore’s IPU. The throughput measured here is the number of images processed per second during training. Each network architecture is ran 60 epochs 20 times on each hardware. The throughput is calculated by:
% \begin{enumerate}
%   \item measure the time taken to process one minibatch
%   \item divided the batch size with the time measured (For this experiment, the default batch size chosen is 128).
% \end{enumerate}

%The result can be seen in Table II.
%TABLE II. 	THROUGHPUT PERFORMANCE OF THE THREE HARDWARE
%	Throughput Metric	Nvidia GTX 1080	Nvidia V100	Graphcore IPU
%SNN-FCN	Average	5287.702	7207.240	46297.169
%	Standard Deviation	124.577	61.026	3622.501
%	Improvement \usepackage{}%	0.0	+36.302	+775.563
%SNN-CNN	Average	3054.903	3883.886	15566.157
%	Standard Deviation	37.826	29.909	1069.288
%	Improvement \%	0.0	+27.136	+409.547

% In section II A, it was mentioned the IPU is optimised for sparse matrix operation, and as seen in Figure 14, the IPU's throughput with sparse matrix as input is significantly higher compared to the GPU.

\section{Outlook and Conclusion}
SNNs and conventional neural networks have overlapping features that can be concurrently optimized, and IPUs have demonstrated promising suitability for most operations that are characteristic of training SNN workloads. We flip the conventional approach to ASIC-driven SNN training by tailoring pre-compiled microcode to a domain-specific accelerator, rather than reconfiguring neuromorphic chips to handle backpropagation approximations on fixed network architectures. Our results show promising performance gains (throughput, TOPS/W, accuracy) can be made by porting the advances made in deep learning accelerators to SNNs. We also indicate the types of hardware optimizations that benefit gradient-based learning in SNNs, such as MIMD processing, functional outlining, and balanced compilations of neuronal and synaptic operators, and how population encoding can be used to better utilize parallelism across both IPUs and GPUs. These features together enable high performance training and inference speeds with IPUs on SNNs.

All code used to generate these results is made openly accessible to enable the research community to accelerate their own custom SNNs on IPUs, and can be installed via PyPi. Population encoding has been integrated into \textit{snnTorch}, with a corresponding interactive notebook that enables users to train population encoded SNNs on both IPUs and GPUs alike.\footnote{URL: https://snntorch.readthedocs.io/en/latest/tutorials/tutorial\_pop.html}

\section*{Conflict of Interest}
A. Titterton, A. Gopiani, and T. Santos are employees at Graphcore. The remaining authors have no conflicts of interest to declare.

% In this paper, we present the advantages of using dedicated hardware to training SNNs. The experiments showed that using hardware that is designed specifically for AI/ML application can improve the training throughput of SNN models while consuming significantly less power, especially when the key operations are able to directly interface with hardware to further improve performances. Not only is the IPU ideal for vanilla SNNs, it also shows how versatile it is by being able to train at high throughput with different spiking dynamics.

\color{black}
% \bibliographystyle{IEEEtran}  

% \begin{thebibliography}{1}
\bibliographystyle{IEEEtran}

% \end{thebibliography}

% \bibliography{refs} 

\end{document}